\documentclass[runningheads]{llncs}

\usepackage{eccvabbrv}

\usepackage{graphicx}
\usepackage{tabularx}
\usepackage{rotating}
\usepackage{booktabs}
\usepackage{makecell}
\usepackage{amssymb}
\usepackage{array}
\usepackage[labelfont=bf]{caption}
\usepackage{longtable}
\usepackage{xcolor}
\definecolor{vermillion}{HTML}{D55E00}
\definecolor{deepblue}{HTML}{0072B2}

\makeatletter
\renewcommand\subparagraph{\@startsection{subparagraph}{5}{\parindent}%
  {3.25ex \@plus1ex \@minus .2ex}%
  {-1em}%
  {\normalfont\normalsize\bfseries}}
\makeatother
\usepackage{titlesec}
\titlespacing*{\section}{0pt}{14pt plus 2pt minus 2pt}{8pt plus 2pt minus 2pt}
\titlespacing*{\subsection}{0pt}{8.8pt plus 2pt minus 2pt}{4.8pt plus 2pt minus 2pt}

\newcolumntype{Y}{>{\raggedright\arraybackslash}X}

\renewcommand\theadgape

\renewcommand{\arraystretch}{1.2}

\usepackage{algorithm}
\usepackage{algpseudocode}
\usepackage{microtype}
\usepackage[accsupp]{axessibility}  

\usepackage[normalem]{ulem}

\usepackage{changes}

\usepackage{hyperref}
\usepackage[nameinlink,noabbrev]{cleveref}

\usepackage{orcidlink}
\usepackage{multirow}

\begin{document}


\title{ChiroEcho: extending automated bat vocalisation classification beyond the learned taxonomy}

\titlerunning{ChiroEcho: extending automated bat vocalisation classification}

\author{Burooj Ghani\inst{1}\orcidlink{0000-0002-0541-3571} \and
Welmoed Eversteijn\inst{1,2}\orcidlink{0009-0003-0036-025X} \and
Milan van Hirtum\inst{1}\orcidlink{0009-0003-5962-9794} \and
Juan Sebastián Cañas\inst{1,3}\orcidlink{0000-0003-0365-5005} \and
Vincent J. Kalkman\inst{1}\orcidlink{0000-0002-1484-7865} \and
Dan Stowell\inst{1,2, 4}\orcidlink{0000-0001-8068-3769} \and 
\\A. Leonie Baier\inst{1}\orcidlink{0000-0002-0327-0378}}

\authorrunning{B. Ghani et al.}

\institute{Naturalis Biodiversity Center \and
Department of Cognitive Science and Artificial Intelligence, Tilburg University \and
People and Nature Lab, University College London \and
Leiden Institute of Advanced Computer Science, Leiden University \\
\email{\{burooj.ghani,leonie.baier\}@naturalis.nl}}

\maketitle

\begin{abstract}
Bats are key indicators of ecosystem health and are protected throughout Europe, making reliable population monitoring a conservation priority. Their cryptic nocturnal lifestyle makes passive acoustic monitoring essential, yet automated identification remains difficult as echolocation calls vary with behaviour and environment and overlap among species. We present a deep learning framework that jointly predicts species and genus and combines genus predictions with geographic species distributions at inference. When only one species of a predicted genus occurs in a region, the framework can resolve species absent from the learned taxonomy. This reframes geographic information as a means of extending, rather than constraining, a classifier’s effective taxonomy. Using recordings spanning 35 European bat species, we evaluate closed-set classification, examine the instability of performance estimates for sparsely represented species, and conduct a controlled held-out proof-of-principle experiment. The rare-species analysis shows how limited evaluation data can obscure species-level performance, while the held-out experiment shows that genus predictions and location can recover labels unavailable to the species head. Geographic resolution extends operational coverage from 35 to 41 of the 48 native European bat species, increasing coverage from 73\% to 85\%. To our knowledge, this is the broadest operational coverage reported for automated European bat classification. More broadly, the bat framework provides proof of principle for resolving unseen fine-grained classes by combining coarse predictions with transparent external constraints.

  \keywords{Bioacoustics \and Deep learning \and Biodiversity monitoring \and Bats \and Ecology \and Ecological reasoning \and Geographical priors}
\end{abstract}


\section{Introduction}
\label{sec:intro}

Bats are ecologically important yet among the least visible inhabitants of European ecosystems. Their nocturnal, cryptic lifestyles~\cite{fentonBatsWorldScience2020} often obscure their roles in insect suppression, nutrient transport, and trophic regulation~\cite{ghanemIncreasingAwarenessEcosystem2012,tuneu-corral2024Bats,sanchez-sotoEcologicalbasedInsightsBat2025}. Many European bat populations have also undergone long-term declines~\cite{browning2021Drivers}, and European bats receive strict legal protection under the EU Habitats Directive~\cite{2025Council}. Their sensitivity to environmental change makes population trends widely used indicators of ecosystem health~\cite{jonesCarpeNoctemImportance2009,russoWeNeedUse2021,siyaBatCaveVulnerability2025}, placing reliable long-term monitoring at the center of biodiversity conservation~\cite{cardinaleBiodiversityLossIts2012}.

These traits make acoustic monitoring indispensable for surveying bat populations~\cite{runkel2021Handbook}. Modern passive acoustic monitoring systems use inexpensive autonomous detectors to collect ultrasonic recordings continuously across unprecedented spatial and temporal scales~\cite{roemerCurrentFrontiersPassive2025}. Converting these recordings into ecological information, however, remains challenging~\cite{russoUseAutomatedIdentification2016,rydell2017Testing,roemerCurrentFrontiersPassive2025}. Whereas automated identification of birds and many other taxa often uses the communication signals, bat classification relies primarily on echolocation calls, whose structure varies not only with species identity but also by flight behaviour, habitat, and foraging context~\cite{denzingerBatGuildsConcept2013}. Consequently, echolocation calls overlap among species and vary within species~\cite{fenton2016Bat}, making reliable identification difficult even for experienced bat workers~\cite{rydell2017Testing}.

Deep learning has substantially improved automated acoustic species recognition~\cite{stowell2022Computational,meramo2026BSGBATS}. However, much of this progress has relied on relatively small, curated datasets suited to controlled comparisons of architectures and training strategies~\cite{bellafkirBatEcholocationCall2022,fundel2023Automatic,alipek2023Efficient}. This focus has accelerated methodological development but often deprioritized taxonomic breadth, limiting the applicability of classifiers for continental-scale monitoring. For bats, broader coverage requires more than acoustic modeling: experts routinely combine acoustic evidence with contextual ecological knowledge~\cite{runkel2021Handbook} during manual identification. Improving automated bat classification need not rely solely on increasingly sophisticated neural architectures; incorporating ecological information at inference offers a complementary path toward greater practical utility.

\begin{figure}[h!]  
   \centering
   \setlength{\abovecaptionskip}{10pt}
   \includegraphics[width=\textwidth]{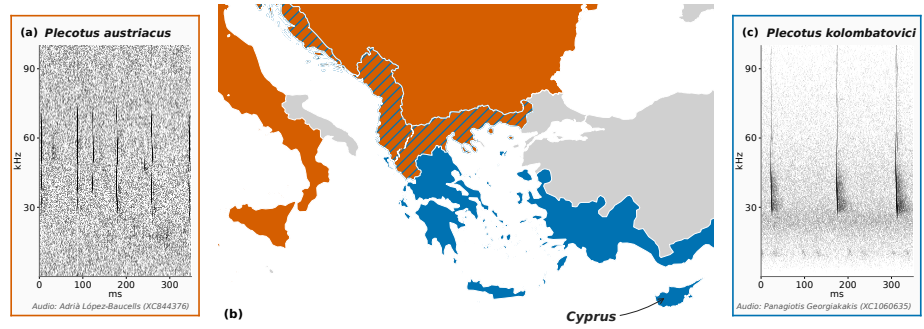}
    \caption{Illustration of inference-stage geographic resolution using the example of \textit{Plecotus kolombatovici}. (a, c) Echolocation call sequences of \textcolor{vermillion}{\textit{P.\ austriacus}} and \textcolor{deepblue}{\textit{P.\ kolombatovici}}, respectively. (b) Approximate Southern European distribution ranges of \textcolor{vermillion}{\textit{P.\ austriacus}} (orange) and \textcolor{deepblue}{\textit{P.\ kolombatovici}} (blue), with overlap shown by orange--blue hatching (ranges based on Dietz et al.~\cite{dietz2024Naturfuehrer}). Owing to insufficient acoustic recordings, \textit{P.\ kolombatovici} is not included in the learned taxonomy. The species head may therefore assign such a recording to an acoustically similar congener class, illustrated here by \textit{P.\ austriacus}. For a recording from Cyprus, a genus-level \textit{Plecotus} prediction above the confidence threshold activates the geographic lookup (\hyperref[tab:geo_constraints]{Table 1}), which resolves the recording to \textit{P.\ kolombatovici} because it is the sole regional representative of the genus. 
    }

  \label{fig:spectograms}
\end{figure}

To demonstrate this principle, we present a framework for automated classification of European bat vocalisations with species- and genus-level classification heads. At inference, each genus prediction is combined with regional species distributions; when only one species of that genus occurs at the recording location, the prediction is resolved unambiguously to that species (\hyperref[fig:spectograms]{Figure 1}). This enables identification of geographically restricted taxa, including species absent from training, thereby extending the framework’s operational taxonomy. To our knowledge, this represents a novel reframing of geographic information in bioacoustic classification: it extends the classifier’s effective taxonomy rather than constraining predictions within a fixed one. More generally, this resolve-when-unambiguous strategy is not tied to acoustic data or geography: any hierarchical classifier could uniquely determine a finer class by combining a reliable coarse prediction with external domain knowledge, such as seasonal occurence or biotic associations. Our bat framework therefore serves as a proof of principle for this broader inference strategy.
\newpage

The main contributions of this work are threefold:

\begin{enumerate}
     
    \item We introduce a joint species- and genus-level classifier that provides explicit genus-level predictions for downstream ecological inference.
    \item We present the broadest operational bat vocalisation classifier for European bats to date, extending coverage from 35 trained species to 41 of the 48 native European bat species\footnote[1]{The EUROBATS Agreement currently lists 55 bat species, of which 48 occur within the territory covered by the latest EU Action Plan for bats~\cite{eurobats2022}. See \Cref{tab:eurobats}.}. 
    
    \item We demonstrate a resolve-when-unambiguous strategy in which geographic priors transform genus predictions into species-level identifications for taxa absent from training, providing a proof of principle for extending hierarchical classifiers through external domain knowledge.
\end{enumerate}
\vspace{10pt}
\newpage
\section{Related work}
\subsection{Automated bioacoustic classification}
Automated bioacoustic classification has undergone a rapid transition from pipelines based on handcrafted acoustic features to deep learning approaches, most commonly operating on time–frequency representations such as spectrograms~\cite{goeau2016lifeclef,kahl2021BirdNET,stowell2022Computational,schwinger2026foundation}. Convolutional and transformer-based architectures now enable large-scale analysis across diverse taxa and recording conditions~\cite{stowell2019automatic,kahl2021BirdNET,hagiwara2023aves,canas2025overview,miron2026avex}. Building on this, transfer learning from large audio corpora has become a key strategy for improving performance in low-resource settings~\cite{dufourq2022passive,kong2020panns,ghani2026twelve}, allowing models to generalize across domains and taxa with limited labelled data~\cite{nolasco2023learning,ghani2025Impact,kather2026bacpipe,hummel2026linear}. Despite these advances, most existing systems remain inherently closed-set, restricting predictions to a predefined taxonomy determined during training~\cite{stowell2022Computational,kath2026novel}. This assumption limits their applicability in real-world monitoring, where novel or previously unseen species may occur, motivating recent interest in open-set and open-world bioacoustic classification frameworks that can better handle taxonomic uncertainty and expansion~\cite{cramer2020chirping,morgan2022open,ko2023open,kather2025clustering}.


\subsection{Automated bat vocalisation classification}

Bat-specific studies have similarly shown that spectrogram-based convolutional networks outperform handcrafted acoustic descriptors~\cite{zualkernan2020tiny,chen2020AutomaticStandardizedProcessing,schwab2023Automated}. Subsequent work has addressed joint detection and classification~\cite{macaodhaGeneralApproachBat2022}, transformer architectures~\cite{bellafkirBatEcholocationCall2022,fundel2023Automatic}, multi-label prediction~\cite{dierckx2022Detection}, and behavioural annotation~\cite{vogelbacherDeepLearningRecognizing2023}. Open-source frameworks including NABat ML~\cite{khalighifar2022NABat}, BatDetect2~\cite{macaodhaGeneralApproachBat2022}, BatSpot~\cite{smeele2026BatSpot}, and BSG-BATS~\cite{meramo2026BSGBATS} increasingly emphasise reproducible deployment and community-driven development for passive acoustic monitoring.
Despite these advances, bat classifiers remain constrained by scarce and heterogeneous annotations. Public repositories are typically long-tailed, combine different recording equipment and protocols, and provide weak recording-level labels rather than strongly annotated, temporally localised calls~\cite{chen2020AutomaticStandardizedProcessing}. Community initiatives have expanded access to recordings and annotation tools~\cite{meramo2026BSGBATS}, while data augmentation and transfer learning can improve generalisation under limited supervision~\cite{fundel2023Automatic,ghani2025Impact}. Nevertheless, representative training data remain the principal bottleneck to broad taxonomic coverage.


\subsection{Ecological knowledge in automated classification}
Ecological knowledge remains underused in automated bat classification. In bioacoustic and fine-grained visual species classification, geographic priors have mainly constrained predictions to locally plausible species within a fixed taxonomy~\cite{kahl2021BirdNET,macaodha2019Presenceonly}. We instead use geography to extend the effective taxonomy, resolving geographically restricted species absent from training when genus and location uniquely determine species identity. To our knowledge, this reframing has not previously been explored in either domain.

\section{Methodology}
\subsection{Dataset}
\label{sec:dataset}
We train and evaluate our classifier using ChirosetEurope\footnote[2]{ChirosetEurope: \url{https://doi.org/10.5281/zenodo.20773226}}~\cite{baier2026CSEzenodo}, a curated collection of European bat vocalisations (echolocation and social calls). The dataset comprises 11,517 recordings of 35 European bat species from 32 countries in the western Palaearctic (\Cref{tab:species_composition}). As expected for an opportunistically aggregated bioacoustic dataset, recordings follow a long-tailed distribution: widespread taxa are substantially better represented than geographically restricted species (\Cref{fig:longtail}). We retain this natural imbalance to reflect operational passive acoustic monitoring conditions. ChirosetEurope includes native full-spectrum ultrasonic and historical time-expansion recordings; metadata record each recording's time-expansion factor (TEF), enabling conversion to a common representation during preprocessing. It provides recording-level (weak) species labels rather than temporally localized call annotations; accordingly, the proposed model is trained with weak supervision. See \hyperref[sec:supp]{Supplementary Material} for details.


\subsection{Audio preprocessing}
\label{sec:preprocessing}
All recordings are converted to a common input representation before feature extraction. Each recording is loaded at its native sampling rate, time-expanded tenfold unless already at TEF = 10, and resampled to $32$\,kHz. Recordings shorter than $20$\,s are tiled to $20$\,s; longer recordings are divided into non-overlapping $20$\,s segments. Trailing remainders of at least $2$\,s are tiled to $20$\,s; shorter remainders are discarded. To represent realistic field conditions, we construct a noise class from bat-free recordings in the open-source audio dataset ECOSoundSet~\cite{funosas2026Finely}. Clips are drawn from two complementary subsets: recordings sampled above $100$\,kHz ($250$--$384$\,kHz), which preserve the ultrasonic range relevant to bat monitoring, and recordings sampled at $96$\,kHz, which provide greater acoustic diversity (birdsong, geophony, anthropophony). The number of anthropogenic sounds is capped to prevent class imbalance. Each $4$\,s source clip undergoes the same time expansion and resampling pipeline, yielding $40$\,s of audio, from which the first $20$\,s are retained without padding. Bat and noise clips therefore have the same input format. The resulting $20$\,s clips, hereafter termed \textit{segments}, serve as classifier inputs. During evaluation, segments are processed independently; for recordings containing multiple segments, per-class scores are max-pooled to produce recording-level predictions.

\subsection{Model architecture}
\label{sec:architecture}

ChiroEcho\footnote[3]{\url{https://github.com/bghani/chiroecho}} uses a shared EfficientNet-B3 acoustic encoder~\cite{pmlr-v97-tan19a} with species and genus classification heads. We evaluate two initializations: ChiroEcho$_{\mathrm{IN}}$ uses standard ImageNet-1k weights~\cite{deng2009imagenet} provided by \texttt{timm}, whereas ChiroEcho$_{\mathrm{Perch}}$ uses weights from Perch~2.0~\cite{vanmerrienboer2025Perch}, an EfficientNet-B3 bioacoustic model trained on over 1.5 million labelled recordings of birds, amphibians, insects, and mammals. To enable end-to-end fine-tuning in PyTorch, we map the released JAX/Flax backbone weights layer by layer to the equivalent \texttt{timm} implementation. Transferring only the convolutional backbone, the Perch linear and prototype-learning classification heads, covering 14,795 species, are discarded. We validate the conversion on matched inputs by comparing intermediate activations at the stem, every MBConv block, and the final convolutional head, obtaining near-exact numerical agreement (cosine similarity $>0.999$) throughout.
\vspace{-8pt}
\paragraph{Input representation.} All models use the  Perch~2.0 log-mel spectrogram frontend~\cite{vanmerrienboer2025Perch}: a 20\,ms window with 10\,ms hop (640/320 segments at 32\,kHz), a 1024-point FFT, HTK-scaled mel filterbanks with 128 bins spanning 60\,Hz--16\,kHz, window-sum-normalized magnitude spectra, and log compression with a $10^{-5}$ floor. Using this frontend for both variants isolates the effect of initialization from differences in input representation. Because ChiroEcho$_{\mathrm{IN}}$'s pretrained stem expects three-channel input, we modify its first convolution to accept single-channel spectrograms rather than replicate the input across three channels.
 \vspace{-8pt}
\paragraph{Genus-level auxiliary head.} To address acoustic ambiguity among closely related species, particularly within genera such as \textit{Myotis}, we add an auxiliary genus-classification alongside the species head. Each head is a single linear layer applied to the shared pooled encoder embedding. Genus labels require no additional annotation and are derived by mapping the 35 training species to 11 genera. The heads are trained jointly using 

\begin{equation}
\mathcal{L} = \mathcal{L}_{\text{species}} + \mathcal{L}_{\text{genus}}
\label{eq:combined_loss}
\end{equation}

Both terms are binary cross-entropy losses on multi-label targets, consistent with our mixup strategy, which produces multi-label rather than soft-interpolated targets. This is a standard multi-task setup rather than a hierarchical or conditional classifier: the heads share the encoder but produce predictions independently, without conditioning or gating. The genus objective provides additional supervision and supports geographic resolution (\hyperref[sec:geo_disambiguation]{Section 3.6}), where genus predictions combined with location metadata can recover certain species absent from or underrepresented in the training data.

\subsection{Training protocol}
\label{sec:training}

\paragraph{Training strategy.} Both ChiroEcho variants are trained end-to-end, with no frozen backbone layers and randomly initialized linear heads after global average pooling. We use AdamW~\cite{loshchilov2017decoupled} with a learning rate and weight decay of $1\times10^{-4}$, a batch size of 32, and no gradient accumulation. The backbone and both heads are optimized jointly using the multi-label binary cross-entropy objective in \hyperref[eq:combined_loss]{Equation 1}. Training uses automatic mixed precision and gradient clipping with a maximum norm of $1.0$. A reduce-on-plateau scheduler (mode \texttt{max}) monitors validation mean average precision (mAP), reducing the learning rate by a factor of $0.1$ after four epochs without improvement, to a minimum of $1\times10^{-6}$. Because species classification is the primary objective, checkpoint selection and early stopping are based on validation species mAP. Training stops after five consecutive epochs without improvement, yielding final checkpoints at epochs 14 for ChiroEcho$_{\mathrm{IN}}$ and 22 for ChiroEcho$_{\mathrm{Perch}}$.

To mitigate class imbalance, examples are sampled with replacement using capped inverse-frequency weights. For class \(c\), with \(n_c\) training examples, the class weight is
\begin{equation}
w_c = \min\{n_c, 1000\}^{-0.5}
\label{eq:sampling_weight}
\end{equation}

The cap assigns equal weights to classes with at least 1,000 examples, focusing oversampling on underrepresented classes. The resulting per-example weights define the weighted random sampler.
\vspace{-8pt}
\paragraph{Data augmentation.} We apply waveform mixup and SpecAugment during training. Mixup is applied with probability $p=0.1$ per batch, using a coefficient $\lambda \sim \text{Beta}(10,10)$ to linearly combine pairs of waveforms. Rather than interpolate their labels using $\lambda$, we retain both source labels as a hard multi-label target, consistent with binary cross-entropy loss. SpecAugment~\cite{park2019specaugment} is applied to every log-mel spectrogram using frequency and time masks with maximum widths of 48 bins and 192 frames, respectively; no frequency or time warping is used.

\subsection{Closed-set species and genus classification}
\label{sec:closed_set}
We first evaluate the model under the standard closed-set assumption: every test recording belongs to one of the trained species classes. This provides a baseline measure of species and genus discrimination ability before considering the open-set extension enabled by geographic resolution (\hyperref[sec:geo_disambiguation]{Section 3.6}), where recordings might belong to species entirely absent from training. Following the max-pooling procedure described above (\hyperref[sec:preprocessing]{Section 3.2}), We report recording-level F1-score, mean average precision (mAP), and area under the receiver operating characteristic curve (AUROC). mAP is the unweighted mean of per-class AP, whereas F1-score and AUROC are averaged using per-class weighting; F1-score is computed by applying a fixed threshold of 0.5 to every class, without threshold tuning, ensuring comparability across settings. Species and genus predictions are evaluated independently using their respective label spaces of 35 species and 11 genera.

\subsection{Geographic resolution}
\label{sec:geo_disambiguation}
\begin{figure}[h!]  
   \centering
   \setlength{\abovecaptionskip}{10pt}
   \includegraphics[width=\columnwidth]{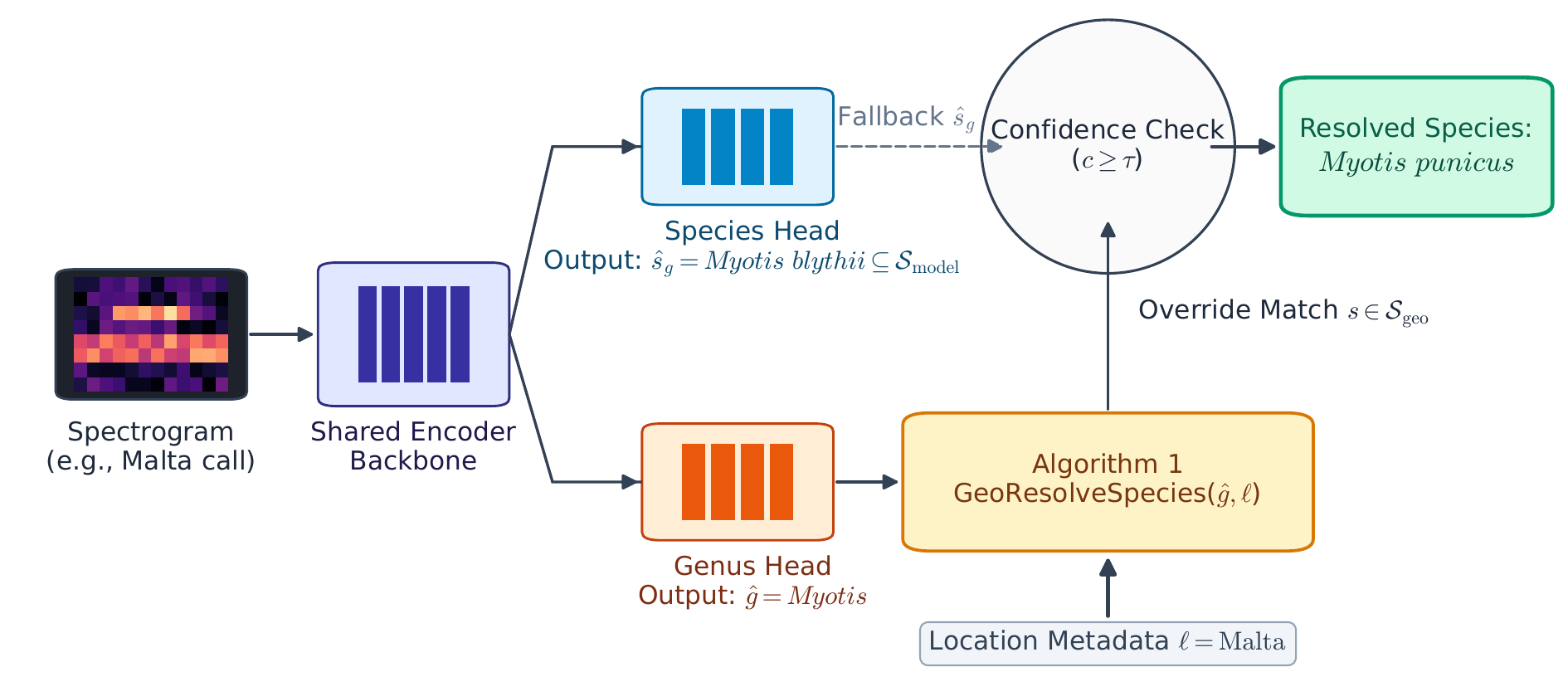}
    \caption{Schematic overview of the ChiroEcho architecture and geographic-resolution step, illustrated for \textit{Myotis punicus} in Malta. The depicted species-head output, \textit{M. blythii}, is illustrative rather than observed and represents a plausible confusion with a closely related species whose echolocation characteristics overlap.}
    
  \label{fig:schematic}
\end{figure}

Prior work has primarily used geographic information to constrain predictions within a fixed taxonomy by favouring locally plausible species. We instead use geography to \emph{extend} the classifier's operational taxonomy beyond the species represented during training. This is possible in regions where only one species of a given genus occurs: once a recording is assigned to that genus, its location uniquely determines species identity (\hyperref[tab:geo_constraints]{Table 1}). For example, a \textit{Myotis} prediction in Malta resolves to \textit{M. punicus} (\hyperref[fig:schematic]{Figure 2}), while a \textit{Plecotus} prediction in Malta resolves to \textit{Pl. gaisleri}.
We implement this inference as a lightweight, rule-based post-processing step combining the genus head (\hyperref[sec:architecture]{Section 3.3}) with recording location metadata. \hyperref[alg:geo_disambiguation_final]{Algorithm 1} is applied independently to each predicted genus, allowing predictions for multiple genera within a recording. For a genus prediction \(\hat{g}\) with confidence \(c \geq \tau\), the rule queries \(\operatorname{GeoResolveSpecies}(\hat{g}, \ell)\), where \(\ell\) is the recording location. The lookup returns a species \(s\) only when that species is the sole representative of genus \(\hat{g}\) in the region; otherwise, it returns \(\varnothing\). When \(s\) is returned, it inherits the genus confidence \(c\) and replaces the species-head predictions \(\hat{s}_g\) for that genus. Predictions for other genera remain unchanged. If \(c < \tau\) or the lookup returns \(\varnothing\), the corresponding species-head predictions are retained (\hyperref[fig:schematic]{Figure 2}).
\begin{algorithm}[h!]
\caption{Geographic resolution rule}
\label{alg:geo_disambiguation_final}
\begin{algorithmic}[1]
\Statex  $S_{\mathrm{model}}$: species represented during training; $S_{\mathrm{geo}}$: species that may be returned by the geographic lookup
\Statex\hspace{-\algorithmicindent}\rule{\linewidth}{0.4pt}
\Require genus prediction $\hat{g}$ with confidence $c$, user-defined threshold $\tau$, recording location $\ell$, scored species-head predictions $\hat{s}_g$ for genus $\hat{g}$, lookup function $\mathrm{GeoResolveSpecies}(\hat{g}, \ell)$ (one-to-one mapping) 
\State $s \gets \emptyset$
\If{$c \geq \tau$}
\State $s \gets \mathrm{GeoResolveSpecies}(\hat{g}, \ell)$
\Comment{$s \in \mathcal{S}_{\text{geo}} \cup \{\emptyset\}$}
\EndIf
\If{$s \neq \emptyset$}
\State \textbf{return} $\{(s,c)\}$
\Comment{\parbox[t]{6.2cm}{geographically resolved single species inherits genus confidence and overrides $\hat{s}_g$}}
\EndIf
\If{$\hat{s}_g \neq \emptyset$}
\State \textbf{return} $\hat{s}_g$
\Comment{fallback to species-head predictions $\hat{s}_g \subseteq \mathcal{S}_{\text{model}}$}
\EndIf
\State \textbf{return} $\emptyset$
\Comment{no valid genus or species prediction for genus $\hat{g}$}
\end{algorithmic}
\end{algorithm}

This mechanism can identify species absent from training, provided that their genus is resolvable and they are its sole regional representative. In the current implementation, the lookup contains 11 region–genus mappings across seven regions and resolves eight species, six of which are absent from training (\hyperref[tab:geo_constraints]{Table 1}). It therefore extends the classifier’s operational taxonomy from 35 to 41 European bat species. The rule requires no additional training and remains transparent and readily extensible as species-distribution knowledge evolves. Its applicability is limited to region–genus pairs that uniquely determine species identity and depends on accurate location and distribution data.
\vspace{-25pt}
\begin{table}[h]
\caption{Region--genus
lookup table used to resolve species $s$ from predicted genus $\hat{g}$ and location
$\ell$ (\hyperref[alg:geo_disambiguation_final]{Algorithm 1}). A genus prediction is
resolved only when a single species of that genus occurs in the listed region. ``In-region data'' indicates that labeled recordings of the resolved species were available from that region. ``Taxonomy extension'' indicates that
the resolved species was absent from the training taxonomy.}
\label{tab:geo_constraints}
\centering
\small

\begin{tabularx}{\columnwidth}{@{}YYYcc@{}}
\toprule
\multicolumn{1}{l}{Genus} &
\multicolumn{1}{l}{Region} &
\multicolumn{1}{l}{Resolved species} &
\multicolumn{1}{l}{\makecell[c]{In-region\\data}} &
\multicolumn{1}{c}{\makecell[c]{Taxonomy\\extension}} \\
\midrule

\textit{Cnephaeus}   & Gibraltar    & \textit{C.\ isabellinus}        &  & \checkmark \\
\textit{Myotis}      & Malta        & \textit{M.\ punicus}            &  & \checkmark \\
\textit{Myotis}      & Pantelleria  & \textit{M.\ punicus}            &  & \checkmark \\
\textit{Nyctalus}    & Azores       & \textit{N.\ azoreum}            &  & \checkmark \\
\textit{Plecotus}    & Malta        & \textit{Pl.\ gaisleri}          &  & \checkmark \\
\textit{Plecotus}    & Pantelleria  & \textit{Pl.\ gaisleri}          &  & \checkmark \\
\textit{Plecotus}    & Cyprus       & \textit{Pl.\ kolombatovici}     &  & \checkmark \\
\textit{Plecotus}    & Canary Is.   & \textit{Pl.\ teneriffae}        &  & \checkmark \\
\midrule
\textit{Pipistrellus} & Pantelleria  & \textit{P.\ kuhlii}             &  &  \\
\textit{Pipistrellus} & Azores       & \textit{P.\ maderensis}         &  &  \\
\textit{Pipistrellus} & Madeira      & \textit{P.\ maderensis}         & \checkmark &  \\
\bottomrule
\end{tabularx}
\end{table}
\vspace{-15pt}

To evaluate whether the rule can recover species outside the learned taxonomy, we trained a separate model with \textit{Pipistrellus maderensis} and \textit{P. kuhlii} excluded from both the training data and the species-head output classes. The held-out recordings from these two species were retained solely for evaluation, making direct prediction by the species head impossible. For \textit{P. maderensis}, we applied the rule using observed recording locations in Madeira. We selected \textit{P. kuhlii} because, aside from \textit{P. maderensis}, it was the only species in the geographic lookup table with sufficient recordings for a quantitative held-out evaluation (n=803). Because none of these recordings originated from Pantelleria, where \textit{P. kuhlii} is the sole representative of \textit{Pipistrellus}, we assigned Pantelleria as their location metadata. This provides a controlled test of the inference mechanism. Together, the two conditions test whether genus predictions and geographic information can extend operational species coverage beyond the learned taxonomy. Results are reported in \hyperref[sec:Geographic resolution]{Section 4.2}.
\vspace{-10pt}
\section{Results}
We first evaluate classification described in \hyperref[sec:closed_set]{Section 3.5}, where all test recordings belong to classes represented during training, and examine performance on sparsely represented species. We then evaluate the geographic resolution strategy using the held-out proof-of-principle experiment described in \hyperref[sec:geo_disambiguation]{Section 3.6}.

\subsection{Closed-set species and genus classification}
\label{sec:closed-set-results}

\hyperref[tab:bat-results1]{Table 2} reports recording-level species and genus performance for the two ChiroEcho variants. ChiroEcho$_{\mathrm{Perch}}$ achieved slightly higher F1-scores at both taxonomic levels, whereas ChiroEcho$_{\mathrm{IN}}$ achieved higher mAP and AUROC, obtaining the best result for four of the six metrics. Genus-level F1-score and mAP exceeded species-level performance for both variants; genus-level AUROC was higher for ChiroEcho$_{\mathrm{IN}}$ and equal to species-level AUROC for ChiroEcho$_{\mathrm{Perch}}$. Overall, the two initialization strategies performed comparably.
\vspace{-15pt}
\begin{table}[h]
    \centering
    \caption{Recording-level closed-set species and genus classification performance of ChiroEcho, trained on 35 European bat species and one noise class with ImageNet-1k or Perch 2.0 initialization. Segment-level predictions are aggregated to recording-level predictions by max pooling before evaluation. Best performance per metric is shown in bold.}
    \label{tab:bat-results1}
    \begin{tabular}{l@{\hspace{12pt}}l@{\hspace{12pt}}ccc@{\hspace{12pt}}ccc}
        \toprule
        \multirow{2}{*}{Model} & \multirow{2}{*}{\shortstack{Weights\\initialization}} & \multicolumn{3}{c}{Species} & \multicolumn{3}{c}{Genus} \\
        \cmidrule(lr){3-5} \cmidrule(lr){6-8}
        & & F1-score & mAP & AUROC & F1-score & mAP & AUROC \\
        \midrule
        ChiroEcho$_{\mathrm{IN}}$ & ImageNet-1k & 0.836          & \textbf{0.694} & \textbf{0.990}          & 0.880         & \textbf{0.923} & \textbf{0.992}          \\
        ChiroEcho$_{\mathrm{Perch}}$ & Perch 2.0 & \textbf{0.845} & 0.672          & 0.988 & \textbf{0.886} & 0.921          & 0.988 \\
        \bottomrule
    \end{tabular}
\end{table}
\vspace{-10pt}

Aggregate metrics can obscure uncertainty for sparsely represented species. \hyperref[tab:low_sample_species]{Table 3} therefore reports AP for the nine species represented by fewer than 10 test-split segments using ChiroEcho$_{\mathrm{IN}}$. Test-split average precision (AP) ranged from 0.001 for \textit{M.\ blythii} to 1.0000 for \textit{M.\ davidii}, although the latter was based on a single test segment. To assess sensitivity to the small sample sizes, we also calculated AP on the combined validation and test splits, increasing the number of segments per species from 1–9 to 2–50. AP changed substantially for several species, including \textit{M. emarginatus} (0.004 to 0.335), \textit{M. capaccinii} (0.501 to 0.817), and \textit{Plecotus austriacus} (0.825 to 0.374), underscoring the uncertainty associated with these sparse evaluations.

\begin{table}[h]
    \centering
    \caption{Classification results for species represented by fewer than 10 test-split segments using ChiroEcho$_{\text{IN}}$. Average precision (AP) is reported for both the test split and the combined validation+test split.}
    \label{tab:low_sample_species}
    \begin{tabular}{lcccc}
        \toprule
        \multirow{2}{*}{Species} & \multicolumn{2}{c}{\# Segments} & \multicolumn{2}{c}{Average precision} \\
        \cmidrule(lr){2-3} \cmidrule(lr){4-5}
        & Test & Val+Test & Test & Val+Test \\
        \midrule
        \textit{Myotis davidii}          & 1 & 2  & 1.000 & 1.000 \\
        \textit{Myotis alcathoe}           & 1 & 11  & 0.019 & 0.013 \\
        \textit{Myotis blythii}           & 2 & 15  & 0.001 & 0.001 \\
        \textit{Myotis mystacinus}          & 2 & 42  & 0.019 & 0.074 \\
        \textit{Myotis emarginatus}          & 4 & 7  & 0.004 & 0.335 \\
        \textit{Myotis bechsteinii}       & 5 & 9  & 0.138 & 0.052 \\
        \textit{Myotis capaccinii}       & 6 & 26  & 0.501 & 0.817 \\
        \textit{Myotis brandtii}         & 9 & 50  & 0.303 & 0.493 \\
        \textit{Plecotus austriacus}      & 9 & 11 & 0.825 & 0.374 \\
    
        \bottomrule
    \end{tabular}
\end{table}

Because genus labels can also be inferred from species predictions, we compared the auxiliary genus head with the genus implied by the top-1 species prediction. The two agreed for 97.74\% of test recordings and disagreed for 2.26\%. Top-1 species accuracy was 91.53\% when the predictions agreed but only 32.81\% when they disagreed (Fisher's exact test, odds ratio $= 22.1$, $p = 1.2 \times 10^{-29}$). Disagreement between the heads therefore provides a strong indicator of species-level prediction error.
\vspace{15pt}
\subsection{Geographic resolution}
\label{sec:Geographic resolution}
We next evaluate the central contribution of this work: extending species recognition beyond the training taxonomy through inference-stage geographic resolution. \hyperref[tab:geo_holdout_results]{Table 4} summarizes the held-out endemic species experiment described in \hyperref[sec:geo_disambiguation]{Section 3.6}: all \textit{Pipistrellus kuhlii} and \textit{P.\ maderensis} recordings and their species-head output classes were removed from the training taxonomy, leaving 33 trained species plus the noise class. All available segments from the held-out species were used for evaluation. 
\vspace{-18pt}
\begin{table}[h!]
    \centering
    \caption{Geographic resolution of two \textit{Pipistrellus} species held out entirely from the training taxonomy and species-head label space. All available segments were used for evaluation. Because neither species has an output class, direct identification by the species head is impossible; the “Top confusion” columns instead report the most frequent segment-level top-1 prediction among the trained classes and its share of segments. Geo-resolved species accuracy is reported at the recording level after max pooling and application of the geographic resolution rule (\hyperref[alg:geo_disambiguation_final]{Algorithm 1}).}

    \label{tab:geo_holdout_results}
    \begin{tabular}{lc@{\hspace{10pt}}lc@{\hspace{10pt}}c}
        \toprule
        \multirow{2}{*}{Species} & \multirow{2}{*}{\# Segments} & \multicolumn{2}{c}{Species head (baseline)} & \multirow{2}{*}{\shortstack{Geo-resolved\\ species accuracy}} \\
        \cmidrule(lr){3-4}
        & & Top confusion & \% & \\
        \midrule
        \textit{P.\ kuhlii}      & 803 & \textit{P.\ nathusii} & 37.4\% & 0.695 \\
        \textit{P.\ maderensis}  & 44  & \textit{P.\ pipistrellus} & 93.2\% & 0.954 \\
        \bottomrule
    \end{tabular}
\end{table}
\newpage
Because neither species had an output class, the species head necessarily predicted incorrect alternatives from the training taxonomy, most often acoustically similar \textit{Pipistrellus} species. Its most frequent segment-level predictions were \textit{P. nathusii} for \textit{P. kuhlii} (300/803 segments; 37.4\%) and \textit{P. pipistrellus} for \textit{P. maderensis} (41/44; 93.2\%). In contrast, combining the genus head with geographic resolution recovered the correct held-out species with recording-level accuracies of 69.5\% and 95.4\%, respectively. Because the lookup is deterministic, successful resolution depends on the genus head identifying \textit{Pipistrellus} above the confidence threshold. These results provide proof of principle that genus-level predictions combined with geographic information can recover species absent from the learned taxonomy.
\section{Discussion}
Overall, our results show that explicit geographic information can complement acoustic classifiers to extend the species coverage beyond the training taxonomy while maintaining strong closed-set performance. In doing so, the framework also broadens the operational taxonomic scope of automated European bat classification. We discuss the implications of these findings below.

We first consider the influence of initialization on the underlying acoustic classifier. ChiroEcho$_{\mathrm{IN}}$, initialized with ImageNet-1k weights, achieved the highest mAP and AUROC at both taxonomic levels, whereas ChiroEcho$_{\mathrm{Perch}}$ achieved slightly higher F1-scores. F1-score depends on a single operating threshold and is sensitive to calibration and class distribution, whereas mAP and AUROC assess discrimination across thresholds. We therefore emphasize mAP and AUROC when comparing the initialization strategies to provide a more robust assessment of the model's discriminative ability. Based on these metrics, ChiroEcho$_{\mathrm{IN}}$ provides the strongest overall acoustic classifier and serves as the default ChiroEcho model in the remaining experiments.

The principal contribution of this work lies in the inference-stage geographic resolution mechanism. The held-out proof-of-principle experiment (\hyperref[tab:geo_holdout_results]{Table 4}) evaluated two species absent from a separately trained 33-species taxonomy, making direct species-head prediction impossible. Rather than failing arbitrarily, the species head consistently confused held-out recordings with acoustically similar congeners, suggesting that the learned embedding captures meaningful taxonomic structure even beyond the trained output classes. Geographic resolution recovered the correct species with recording-level accuracies of 69.5\% and 95.4\%, respectively, without requiring any species-specific training data. This mechanism is inherently limited to taxa for which geographic occurrence uniquely determines species identity (\hyperref[tab:geo_constraints]{Table 1}) and therefore cannot be applied to many species with limited acoustic training data. For applicable species, however, it demonstrates how ecological knowledge can complement learned acoustic representations. Because the rule operates only at inference, updated species distributions, taxonomic revisions, or newly recognized endemic populations can be incorporated without retraining the acoustic model.

Outside geographically resolvable cases, performance remains constrained by the availability of acoustic training and evaluation data. \hyperref[tab:low_sample_species]{Table 3} illustrates the instability of per-species evaluation under severe class imbalance. With as few as one or two test segments, average precision is dominated by individual predictions rather than reflecting general model behavior: a single correct or incorrect classification can shift the reported score from 0 to 1. The perfect AP obtained for \textit{M.\ davidii} (1 test segment) should therefore not be interpreted as evidence of reliable performance on this species, but rather as an artifact of the evaluation size. Combining the validation and test splits provides a broader estimate, but several species remain represented by only 2–11 segments, and their AP values vary substantially between splits. We view this as an inherent constraint of working with long-tailed, data-scarce taxa in bioacoustic monitoring, rather than a shortcoming specific to our model, and note that the low training segment counts for these species (10--40 segments) likely contribute directly to their inconsistent performance. This presents a practical challenge for conservation, as many species for which reliable monitoring is most important are also among those with the scarcest acoustic training data, creating a self-reinforcing cycle in which limited data availability constrains classifier performance precisely where automated monitoring could provide the greatest benefit. Targeted collection of recordings from underrepresented species should therefore remain a priority.

A second limitation is domain shift. Both training and evaluation use curated repository recordings, whereas operational passive acoustic monitoring encounters more complex soundscapes with overlapping vocalisations, environmental noise, and recording artefacts. Evaluating the proposed framework on long-duration, uncurated recordings is therefore an important next step toward deployment.

Beyond its role in geographic resolution, the auxiliary genus head provides informative genus-level predictions when species-level discrimination is uncertain. Genus-level F1-score and mAP exceeded species-level values for both initialization strategies; genus-level AUROC was higher for ChiroEcho$_{\mathrm{IN}}$ and equal for ChiroEcho$_{\mathrm{Perch}}$. This pattern is expected as distinguishing among genera is generally easier than distinguishing among closely related species with overlapping vocal characteristics. The genus head therefore supports graceful degradation: when a species prediction is unreliable, the system can report a genus and confidence score, narrowing the set of plausible species for expert review. This is particularly relevant for rare species such as those in \hyperref[tab:low_sample_species]{Table 3}, whose species-level training data may remain scarce.

The genus implied by the top-1 species prediction agreed with the dedicated genus head for 97.74\% of test recordings, indicating that an explicit head is not strictly necessary to obtain genus-level predictions. We nevertheless retain it as an architectural choice, for two reasons. First, disagreement between the heads provides a useful uncertainty signal unavailable to a purely species-derived genus score: species accuracy fell from 91.53\% on agreeing recordings to 32.81\% on disagreeing recordings. Second, the dedicated head provides a directly thresholdable genus output without requiring an additional aggregation rule over multi-label species outputs, at negligible additional computational cost.
\vspace{-10pt}
\section{Conclusion}
We presented a deep learning framework for automated classification of European bat vocalisations that combines species- and genus-level prediction with ecological domain knowledge. Trained on 35 species, the proposed framework extends operational coverage to 41 of the 48 native European bat species by combining learned genus predictions with an inference-stage geographic rule. This increases operational taxonomic coverage from 73\% to 85\% without additional species-specific training examples and, to our knowledge, represents the broadest operational coverage reported for automated European bat classification.

More broadly, this study illustrates how transparent domain knowledge can extend a learned classifier without increasing architectural complexity. Although we demonstrate the resolve-when-unambiguous principle for genus-to-species acoustic classification in European bats, it could extend the effective taxonomy of other hierarchical classifiers whenever a reliable coarse prediction and an external ecological constraint uniquely determine a fine-grained class. Such constraints might include regional checklists, host associations, or seasonal occurrence, with potential applications to other acoustically monitored taxa, camera-trap imagery, and eDNA surveys. We therefore view our bat pipeline as one instance of a broader strategy for taxonomic resolution through auxiliary ecological information.

Future work will focus on two complementary directions. First, the proposed framework should be evaluated on uncurated passive acoustic monitoring soundscapes to assess its robustness to the domain shift between curated reference recordings and real-world monitoring data. Strongly annotated soundscape datasets will be essential both for this evaluation and for further improving the underlying acoustic classifier as they become available. Second, we will extend the ecological inference framework by incorporating additional forms of ecological domain knowledge. An important next step will be to provide geographically plausible candidate species that are absent from the learned taxonomy, allowing users to distinguish between species that can be identified directly by the classifier and those that remain ecologically plausible but are not yet represented in the training data. This would improve the transparency and practical utility of automated biodiversity monitoring systems while accommodating future taxonomic revisions, refined species distributions, and newly available training data. Rather than treating ecological knowledge as external to machine learning, we envision biodiversity monitoring systems in which learned acoustic representations and explicit ecological reasoning operate as complementary components of a unified inference pipeline.
\newpage
\section*{CRediT authorship contribution statement}
BG: Conceptualization, Methodology, Software, Validation, Formal analysis, Investigation, Writing – original draft, Writing – review \& editing. WE: Software. MH: Data curation. JSC: Conceptualization, Writing – review \& editing, Visualization. VJK: Methodology, Supervision, Project administration, Funding acquisition. DS: Methodology, Supervision, Project administration, Funding acquisition. ALB: Conceptualization, Methodology, Validation, Investigation, Data curation, Writing – original draft, Writing – review \& editing, Visualization, Supervision.

\section*{Acknowledgements}
This work was supported by funding received via the Horizon Europe projects TETTRIs (grant no. 101081903, supporting work by BG and VJK) and MAMBO (grant no. 101060639, supporting work by BG, MH, VJK, DS, and ALB) as well as the UK Research and Innovation (UKRI) Horizon Europe Guarantee (Grant EP/Y033299/1), as part of the Marie Skłodowska-Curie Actions Doctoral Networks Programme, BioacAI (grant no. 101116715, supporting work by JSC). The views and opinions expressed are, however, those of the authors only and do not necessarily reflect those of the European Union or the European Research Executive Agency (REA). The funders had no role in study design, data collection and analysis, decision to publish, or preparation of the manuscript.

\bibliographystyle{splncs04}
\bibliography{main}

\clearpage
\section*{Supplementary Material}
\label{sec:supp}

\setcounter{table}{0}
\renewcommand{\thetable}{S\arabic{table}}

\setcounter{figure}{0}
\renewcommand{\thefigure}{S\arabic{figure}}

\setcounter{table}{0}
\setcounter{figure}{0}

\renewcommand{\thetable}{S\arabic{table}}
\renewcommand{\thefigure}{S\arabic{figure}}

\renewcommand{\theHtable}{supp.table.\arabic{table}}
\renewcommand{\theHfigure}{supp.figure.\arabic{figure}}

\vspace{10pt}
\subsection*{ChirosetEurope dataset}
\phantomsection
\label{sec:chiroset_dataset}
\vspace{5pt}
\paragraph{Taxonomic coverage.} To contextualize the taxonomic coverage of ChirosetEurope, we compared its contents with the 55 bat species currently listed under the EUROBATS Agreement~\cite{eurobats2022}. The Agreement, however, has a broader geographic scope than the latest EU Action Plan for Bats (2018--2024)\cite{barovaStreit2018}, encompassing non-EU European states as well as parts of North Africa and the Middle East. Seven of the 55 species have no documented occurrence within the area covered by the Action Plan, resulting in a reference set of 48 species. This updated total exceeds the 45 species originally recognized by the Action Plan because it incorporates subsequent taxonomic revisions adopted by EUROBATS~\cite{barovaStreit2018,eurobats2022}. ChirosetEurope currently represents 35 of these 48 species (\hyperref[tab:eurobats]{Table S1}).

\begin{sidewaysfigure}[p]
\centering
\includegraphics[width=\linewidth,keepaspectratio]{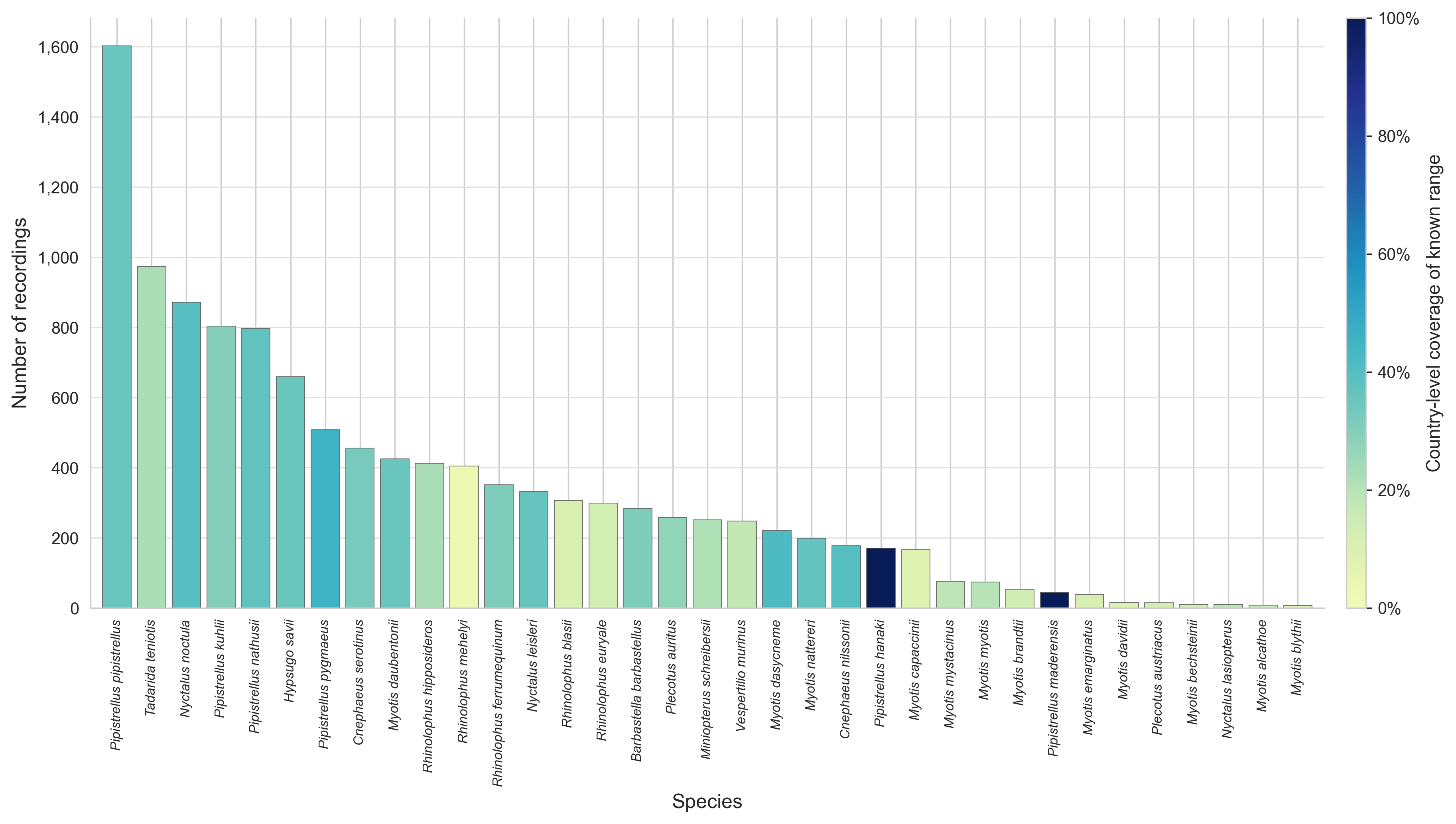}
\caption{\textbf{Recording abundance and relative geographic coverage across the 35 species in ChirosetEurope.} Species are ordered by decreasing recording count, represented by bar height. Bar colour indicates the proportion of each species' reported EUROBATS range states represented by at least one recording, with darker shades indicating greater coverage. Percentages are calculated against the full set of reported range states, including neighbouring countries and range states not represented in ChirosetEurope.}
\label{fig:longtail}
\end{sidewaysfigure}

\paragraph{Composition.} Building on the reference framework described above, \hyperref[tab:species_composition]{Table S2} reports the per-species composition of ChirosetEurope and its operational taxonomic coverage. \hyperref[fig:longtail]{Figure S1} shows recording abundance and relative geographic coverage, highlighting the long-tailed distribution arising from differences in species availability and recording effort. These summaries describe bat recordings only and exclude the separately constructed noise class.

To minimise data leakage, recordings were partitioned using session-level metadata grouping keys. Sessions comprised recordings from the same biological night and approximate location, defined by coordinates rounded to three decimal places (approximately 100~m); recordings made before 12:00 were assigned to the preceding night. Where coordinates were unavailable, recordings were grouped by biological night, recordist, and country. All recordings within a group were assigned to the same split.

ChirosetEurope contains native full-spectrum and historical time-expanded recordings. The time-expansion factor retained in the metadata allows both formats to be transformed to a common input representation. Heterodyne and frequency-division recordings were excluded because these methods irreversibly alter the original ultrasonic signal.

\par
\newpage

\begingroup
\small

\begin{longtable}{@{}lrrl@{}}

\caption{The 55 bat taxa listed under the EUROBATS Agreement on the
Conservation of Populations of European Bats~\cite{eurobats2022},
presented using current nomenclature, and their representation in
ChirosetEurope (`CSE`). `No.` indicates the number of EUROBATS Range States ($n=60$) in which a species is listed as present. Species marked with $^{*}$ have no documented occurrence
within the territory covered by the latest EU Action Plan for Bats
(2018--2024)~\cite{barovaStreit2018}. Data current as of July 2026.}
\label{tab:eurobats}\\

\toprule
Scientific name & CSE & No. & Distribution within EUROBATS area \\
\midrule
\endfirsthead

\multicolumn{4}{@{}l}{%
  \tablename~\thetable\ continued from previous page}\\
\toprule
Scientific name & CSE & No. & Distribution within EUROBATS area \\
\midrule
\endhead

\midrule
\multicolumn{4}{r@{}}{\footnotesize Continued on next page}\\
\endfoot

\bottomrule
\endlastfoot

\textit{Barbastella barbastellus} & \checkmark & 45 & Widespread across Europe and adjacent western Asia \\
\textit{Barbastella caspica}$^{*}$ & & 4 & Caucasus and Caspian region \\
\textit{Cnephaeus anatolicus}$^{*}$ & & 5 & Eastern Mediterranean \\
\textit{Cnephaeus isabellinus} & & 6 & Iberia and North Africa \\
\textit{Cnephaeus nilssonii} & \checkmark & 30 & Northern, central and eastern Europe \\
\textit{Cnephaeus ognevi} & & 6 & Caucasus and Middle East \\
\textit{Cnephaeus serotinus} & \checkmark & 53 & Widespread across Europe, North Africa and western Asia \\
\textit{Hypsugo savii} & \checkmark & 40 & Southern and central Europe, North Africa and western Asia \\
\textit{Miniopterus pallidus}$^{*}$ & & 5 & Caucasus, Turkey and western Asia \\
\textit{Miniopterus schreibersii} & \checkmark & 28 & Southern Europe, the Balkans and the Mediterranean region \\
\textit{Myotis alcathoe} & \checkmark & 31 & Central, southern and eastern Europe \\
\textit{Myotis bechsteinii} & \checkmark & 31 & Western, central and southeastern Europe \\
\textit{Myotis blythii/oxygnathus} & \checkmark & 40 & Southern and central Europe, North Africa and western Asia \\
\textit{Myotis brandtii} & \checkmark & 37 & Northern, central and eastern Europe \\
\textit{Myotis capaccinii} & \checkmark & 24 & Mediterranean region and Middle East \\
\textit{Myotis crypticus} & & 8 & Southwestern Europe \\
\textit{Myotis dasycneme} & \checkmark & 24 & Northern, central and eastern Europe \\
\textit{Myotis daubentonii} & \checkmark & 42 & Widespread across Europe and adjacent western Asia \\
\textit{Myotis davidii} & \checkmark & 11 & Eastern Europe, the Caucasus and western Asia \\
\textit{Myotis emarginatus} & \checkmark & 45 & Southern and central Europe, North Africa and western Asia \\
\textit{Myotis escalerai} & & 3 & France and the Iberian Peninsula \\
\textit{Myotis hoveli} & & 4 & Eastern Mediterranean \\
\textit{Myotis myotis} & \checkmark & 35 & Western, central and southeastern Europe \\
\textit{Myotis mystacinus} & \checkmark & 43 & Widespread across Europe and adjacent western Asia \\
\textit{Myotis nattereri} & \checkmark & 33 & Western, central and northern Europe \\
\textit{Myotis punicus} & & 7 & Western Mediterranean and North Africa \\
\textit{Myotis schaubi}$^{*}$ & & 2 & Armenia and Iran \\
\textit{Myotis tschuliensis}$^{*}$ & & 5 & Caucasus and Turkey \\
\textit{Nyctalus azoreum} & & 1 & Azores (Portugal) \\
\textit{Nyctalus lasiopterus} & \checkmark & 27 & Southern, central and eastern Europe \\
\textit{Nyctalus leisleri} & \checkmark & 47 & Widespread across Europe and adjacent regions \\
\textit{Nyctalus noctula} & \checkmark & 48 & Widespread across Europe and western Asia \\
\textit{Otonycteris hemprichii}$^{*}$ & & 11 & North Africa and Middle East \\
\textit{Pipistrellus hanaki/creticus} & \checkmark & 1 & Greece \\
\textit{Pipistrellus kuhlii} & \checkmark & 44 & Southern and central Europe, North Africa and western Asia \\
\textit{Pipistrellus maderensis} & \checkmark & 2 & Madeira and the Canary Islands \\
\textit{Pipistrellus nathusii} & \checkmark & 43 & Widespread across Europe and western Asia \\
\textit{Pipistrellus pipistrellus} & \checkmark & 53 & Widespread across the EUROBATS area \\
\textit{Pipistrellus pygmaeus} & \checkmark & 44 & Widespread across Europe and western Asia \\
\textit{Plecotus auritus} & \checkmark & 47 & Widespread across Europe and adjacent western Asia \\
\textit{Plecotus austriacus} & \checkmark & 33 & Western, central and southern Europe \\
\textit{Plecotus gaisleri} & & 6 & Central Mediterranean and North Africa \\
\textit{Plecotus kolombatovici} & & 8 & Adriatic, Balkan and eastern Mediterranean region \\
\textit{Plecotus macrobullaris} & & 22 & Mountain regions of southern Europe and western Asia \\
\textit{Plecotus sardus} & & 1 & Sardinia (Italy) \\
\textit{Plecotus teneriffae} & & 1 & Canary Islands (Spain) \\
\textit{Rhinolophus blasii} & \checkmark & 20 & Balkans, eastern Mediterranean and western Asia \\
\textit{Rhinolophus euryale} & \checkmark & 32 & Southern Europe, North Africa and western Asia \\
\textit{Rhinolophus ferrumequinum} & \checkmark & 45 & Widespread across southern Europe and western Asia \\
\textit{Rhinolophus hipposideros} & \checkmark & 50 & Widespread across Europe, North Africa and western Asia \\
\textit{Rhinolophus mehelyi} & \checkmark & 27 & Mediterranean region, the Balkans and western Asia \\
\textit{Rousettus aegyptiacus} & & 9 & Eastern Mediterranean and Middle East \\
\textit{Tadarida teniotis} & \checkmark & 36 & Mediterranean region and western Asia \\
\textit{Taphozous nudiventris}$^{*}$ & & 10 & North Africa and Middle East \\
\textit{Vespertilio murinus} & \checkmark & 40 & Central, northern and eastern Europe \\

\end{longtable}
\endgroup


\begin{table}[H]
\centering
\caption{Per-species composition and operational taxonomic coverage of $ChiroEcho$. For the 35 species represented in ChirosetEurope, the table reports recording counts by split, total audio duration, and the number of recording countries. The lower block lists six additional species supported in $ChiroEcho$ through geographic resolution and the regions in which genus and location uniquely determine species identity. Dataset statistics are not applicable to these species because they are outside the learned taxonomy.}
\vspace{10pt}
\label{tab:species_composition}

\scriptsize
\setlength{\tabcolsep}{5pt}
\renewcommand{\arraystretch}{0.95}

\begin{tabular}{@{}lrrrrrr@{}}
\toprule
\textbf{Learned species} & \textbf{Train} & \textbf{Val} & \textbf{Test} & \textbf{Total} & \textbf{Duration} & \textbf{Countries} \\
\midrule
\textit{Barbastella barbastellus}
  & 170 & 77 & 37 & 284 & 00:46:29 & 14 \\
\midrule
\textit{Cnephaeus nilssonii}
  & 79 & 54 & 44 & 177 & 00:34:44 & 12 \\
\textit{Cnephaeus serotinus}
  & 253 & 123 & 79 & 455 & 01:51:08 & 17 \\
\midrule
\textit{Hypsugo savii}
  & 399 & 100 & 160 & 659 & 01:00:08 & 14 \\
\midrule
\textit{Miniopterus schreibersii}
  & 112 & 109 & 30 & 251 & 00:40:56 & 6 \\
\midrule
\textit{Myotis alcathoe}
  & 5 & 2 & 1 & 8 & 00:00:45 & 5 \\
\textit{Myotis bechsteinii}
  & 6 & 1 & 3 & 10 & 00:01:05 & 4 \\
\textit{Myotis blythii}
  & 3 & 3 & 1 & 7 & 00:00:28 & 3 \\
\textit{Myotis brandtii}
  & 38 & 12 & 3 & 53 & 00:12:11 & 5 \\
\textit{Myotis capaccinii}
  & 152 & 12 & 2 & 166 & 00:19:41 & 2 \\
\textit{Myotis dasycneme}
  & 118 & 63 & 39 & 220 & 00:48:03 & 10 \\
\textit{Myotis daubentonii}
  & 254 & 81 & 90 & 425 & 01:49:22 & 15 \\
\textit{Myotis davidii}
  & 14 & 1 & 1 & 16 & 00:00:30 & 1 \\
\textit{Myotis emarginatus}
  & 35 & 1 & 2 & 38 & 00:08:20 & 5 \\
\textit{Myotis myotis}
  & 66 & 5 & 2 & 73 & 00:23:35 & 7 \\
\textit{Myotis mystacinus}
  & 67 & 8 & 1 & 76 & 00:23:30 & 8 \\
\textit{Myotis nattereri}
  & 119 & 46 & 33 & 198 & 00:44:48 & 12 \\
\midrule
\textit{Nyctalus lasiopterus}
  & 7 & 1 & 2 & 10 & 00:03:20 & 5 \\
\textit{Nyctalus leisleri}
  & 235 & 52 & 45 & 332 & 01:12:24 & 17 \\
\textit{Nyctalus noctula}
  & 518 & 186 & 167 & 871 & 03:21:40 & 19 \\
\midrule
\textit{Pipistrellus hanaki}
  & 76 & 35 & 59 & 170 & 00:13:34 & 1 \\
\textit{Pipistrellus kuhlii}
  & 462 & 77 & 264 & 803 & 02:05:05 & 13 \\
\textit{Pipistrellus maderensis}
  & 22 & 5 & 17 & 44 & 00:16:27 & 2 \\
\textit{Pipistrellus nathusii}
  & 426 & 154 & 216 & 796 & 03:58:11 & 16 \\
\textit{Pipistrellus pipistrellus}
  & 1048 & 344 & 210 & 1602 & 04:56:23 & 19 \\
\textit{Pipistrellus pygmaeus}
  & 297 & 114 & 97 & 508 & 01:26:27 & 20 \\
\midrule
\textit{Plecotus auritus}
  & 158 & 40 & 60 & 258 & 00:27:58 & 13 \\
\textit{Plecotus austriacus}
  & 8 & 1 & 5 & 14 & 00:00:28 & 4 \\
\midrule
\textit{Rhinolophus blasii}
  & 135 & 124 & 47 & 306 & 01:01:27 & 2 \\
\textit{Rhinolophus euryale}
  & 210 & 62 & 27 & 299 & 00:17:30 & 4 \\
\textit{Rhinolophus ferrumequinum}
  & 206 & 109 & 36 & 351 & 00:26:36 & 14 \\
\textit{Rhinolophus hipposideros}
  & 283 & 61 & 68 & 412 & 00:26:17 & 11 \\
\textit{Rhinolophus mehelyi}
  & 269 & 36 & 99 & 404 & 00:20:02 & 1 \\
\midrule
\textit{Tadarida teniotis}
  & 497 & 190 & 287 & 974 & 01:42:04 & 8 \\
\midrule
\textit{Vespertilio murinus}
  & 136 & 44 & 67 & 247 & 00:57:11 & 7 \\
\midrule
\addlinespace
\midrule
\textbf{Geo-resolved species} &
\multicolumn{6}{l}{\textbf{Applicable region(s)}} \\
\midrule
\textit{Cnephaeus isabellinus} &
\multicolumn{6}{l}{Gibraltar} \\
\textit{Myotis punicus} &
\multicolumn{6}{l}{Malta, Pantelleria} \\
\textit{Nyctalus azoreum} &
\multicolumn{6}{l}{Azores} \\
\textit{Plecotus gaisleri} &
\multicolumn{6}{l}{Malta, Pantelleria} \\
\textit{Plecotus kolombatovici} &
\multicolumn{6}{l}{Cyprus} \\
\textit{Plecotus teneriffae} &
\multicolumn{6}{l}{Canary Islands} \\
\bottomrule
\end{tabular}
\end{table}

\end{document}